\documentclass[10pt,twocolumn,letterpaper]{article}

\usepackage[pagenumbers]{cvpr}
\usepackage{multirow}
\usepackage{booktabs}
\usepackage{float}
\usepackage[table]{xcolor}
\usepackage{array}
\usepackage[accsupp]{axessibility}  % Improves PDF readability for those with disabilities.
\usepackage[table]{xcolor}

\definecolor{lightestorange}{RGB}{255,245,235}
\definecolor{lightorange}{RGB}{255,232,214}
\definecolor{mediumorange}{RGB}{255,214,186}
\definecolor{lighttextgray}{RGB}{160,160,160}

\usepackage{placeins}

\definecolor{cvprblue}{rgb}{0.21,0.49,0.74}
\definecolor{lightestgray}{gray}{1}  % Base column
\definecolor{mediumgray}{gray}{0.91}    % +Pose column
\definecolor{darkgray}{gray}{0.91}      % +All column
\definecolor{mygray}{gray}{0.92}
\definecolor{posecol}{RGB}{255,140,0}   % +Pose  (#FF8C00)
\definecolor{allcol}{RGB}{0,136,255}    % +All   (#0088FF)

\usepackage[pagebackref,breaklinks,colorlinks,allcolors=cvprblue]{hyperref}

\def\paperID{*****} % *** Enter the Paper ID here
\def\confName{CVPR}
\def\confYear{2026}

\title{MARIO: Motion-Augmented Real-Time Multi-Sensor Inertial Odometry}

\author{
Yiquan Li\textsuperscript{1,*} \quad
Taeyoung Yeon\textsuperscript{1,*} \quad
Chenfeng Gao\textsuperscript{1} \quad
Vasco Xu\textsuperscript{2} \quad
Xuanyou Liu\textsuperscript{1} \quad
Karan Ahuja\textsuperscript{1} \\
\textsuperscript{1}Northwestern University \quad
\textsuperscript{2}University of Chicago \\
{\small \textsuperscript{*}Equal contribution}
}

\begin{document}
\maketitle
\begin{abstract}

Inertial odometry (IO) using only Inertial Measurement Units (IMUs) provides a lightweight solution for human motion tracking in augmented reality (AR) and wearable devices. Recent learning-based IO methods have improved the generalizability of inertial localization via large-scale pretraining on human motion datasets. Unfortunately, these approaches remain prone to drift and noise because they fail to capture human motion dynamics, especially on daily activity datasets such as Nymeria. In contrast, we propose to ground inertial odometry in human kinematics through a learned IMU-inferred pose prior that promotes the propagation of physically consistent motion constraints. We integrate our pose prior into existing IO architectures and reduce positional drift by up to 36\% in the challenging Nymeria dataset (5x larger than prior works). We further showcase improved long-term performance by developing a sensor-fusion framework that incorporates auxiliary signals from other lightweight sensors such as the magnetometer, barometer, and secondary IMU already available on commercial AR glasses. With our fusion strategy, drift is reduced to 42\%, improving robustness and generalization across diverse motion conditions. Together, our results establish a new paradigm for inertial and lightweight odometry, unifying human motion kinematics with multimodal sensing, setting a new benchmark for accurate and robust camera-less human tracking.  Our website is available at \url{https://spice-lab.org/projects/MARIO/}.

\end{abstract}    
\section{Introduction}
\begin{figure}
    \centering
    \includegraphics[width=\linewidth]{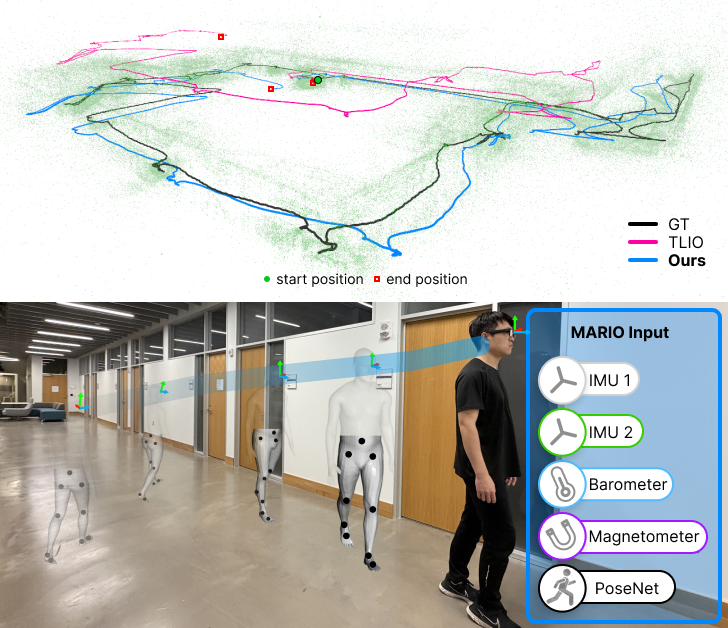}
    \caption{We propose MARIO, an inertial odometry framework that builds on a single-IMU model, grounds motion in lower-body pose predicted by a pretrained PoseNet, and enhances robustness through sensor fusion with a secondary IMU, barometer, and magnetometer.}
    \label{teaser}
\end{figure}

Inertial Measurement Units (IMUs) are compact and low-cost sensors that measure linear acceleration and angular velocity. They are widely used in computer vision, robotics, and extended reality (XR) for motion tracking, operating reliably under challenging conditions with a low power profile and lightweight form-factor. Their small form factor and low power consumption make them a core sensing modality for wearable and mobile devices.

However, estimating position from IMU signals involves integrating noisy acceleration and angular velocity, which leads to cumulative drift. Visual–inertial odometry (VIO) methods \cite{Qin_2018, clark2017vinetvisualinertialodometrysequencetosequence} mitigate this drift by fusing IMU signals with image features from cameras. While effective, visual methods degrade in high-speed motion or visually challenging conditions such as low light, motion blur, or lack of visual features. Moreover, cameras introduce power, privacy, and environmental limitations, making IMU-only odometry increasingly attractive for lightweight AR and wearable devices.

Recently, learning-based inertial odometry (IO) methods have emerged that mitigate drift through data-driven motion priors learned directly from IMU signals \cite{yan2019roninrobustneuralinertial, ma2024nymeriamassivecollectionmultimodal, jayanth2024eqnio, qiu2025airiolearninginertialodometry}. Despite these advances, most models rely on a single IMU and remain prone to drift and noise, limiting their robustness in tracking scenarios that involve complex human motions and diverse environments.

Our key insight is that on-body IMU tracking captures human kinematics, rather than implicitly integrating accelerometer and gyroscope measurements over time. In this work, we explicitly ground inertial odometry in human kinematics through a learned IMU-inferred pose prior, thereby strengthening motion constraints. We propose PoseNet, which predicts full-body pose in the SMPL body model~\cite{loper2023smpl} from a single head-mounted IMU. The learned pose prior serves as a spatio-temporal kinematic anchor, injecting physically meaningful structure into motion estimation and substantially reducing translation error and drift when integrated into existing inertial odometry architectures.

Furthermore, on-body devices such as AR glasses provide readily available, low-power sensors, including a magnetometer, barometer, and dual IMUs. However, the benefits of incorporating these sensors into neural inertial odometry models remain largely unexplored. To further enhance performance, we introduce a Multi-Sensor Fusion Module that incorporates these auxiliary signals, which are already available on commercial AR glasses such as Meta's Aria~\cite{engel2023aria}. These complementary cues provide more accurate heading and elevation information, improving robustness and generalization across diverse motion patterns.

We demonstrate the generality of our framework by integrating it with four state-of-the-art inertial odometry methods and conducting comprehensive evaluations on the Nymeria \cite{ma2024nymeriamassivecollectionmultimodal} (300 hrs, 297.87m avg. trajectory length), Aria Everyday Activities (7.3 hrs, 29.35m avg. trajectory length) \cite{lv2024ariaeverydayactivitiesdataset}, and TLIO \cite{Liu_2020} (60 hrs, 119.78m avg. trajectory length) datasets. Quantitative and qualitative analyses show that our approach consistently reduces drift and lowers translation error across all benchmarks, confirming its robustness and generalization for human-centric motion tracking.

Our key contributions are:
\begin{itemize}
\item We introduce a human-pose-grounded inertial odometry framework that injects a learnt IMU-inferred kinematics prior into our odometry model, improving stability and reducing drift.
\item We demonstrate a sensor-fusion framework that integrates barometer, magnetometer, and dual-IMU signals to provide robust odometry, substantially improving vertical and heading cues.
\item We integrate our modules into existing IO architectures (TLIO \cite{Liu_2020}, AirIO \cite{qiu2025airiolearninginertialodometry}, RoNIN \cite{yan2019roninrobustneuralinertial} and EQNIO \cite{jayanth2024eqnio}) and benchmark on Nymeria, Aria Everyday, and TLIO datasets demonstrating consistent improvements across datasets and metrics.
\end{itemize}

\section{Related Work}

\subsection{Inertial Odometry}
Inertial odometry (IO) estimates a device's 6-DoF pose (position and orientation) from IMU signals. Orientation is obtained by integrating gyroscope readings, while double integration of linear acceleration yields position. However, noise and bias in low-cost IMUs accumulate during integration, causing drift over time. Heuristic methods such as pedestrian dead reckoning (PDR) exploit motion regularities (e.g., step detection and stride estimation) but rely on hand-tuned assumptions and degrade in unconstrained motion \cite{jimenez2009pdr,5649300}.

Learning-based IO replaces such heuristics with data-driven motion priors that infer displacement or velocity directly from IMU sequences. RIDI \cite{yan2018ridi} regresses short-term velocity to correct low-frequency drift, while IONet \cite{chen2018ionet} and RoNIN \cite{yan2019roninrobustneuralinertial} integrate learned velocity estimates for accurate 2D trajectory reconstruction. RoNIN further introduces a heading-agnostic coordinate frame (HACF) aligning gravity with the vertical axis. TLIO \cite{Liu_2020} extends this by combining a learned displacement estimation with an Extended Kalman Filter (EKF) to jointly refine position, orientation, and bias. IDOL \cite{sun2021idol} explicitly estimates orientation from magnetometer readings and regresses translation with a bi-directional LSTM, improving robustness and reducing long-term drift.

To improve generalization, recent works incorporate structural and equivariant priors. AirIO \cite{qiu2025airiolearninginertialodometry} shows that retaining IMU data in the body frame, rather than transforming it to the global frame, improves drone odometry. RIO \cite{cao2022rio} leverages rotational equivariance as a self-supervised signal, and EqNIO \cite{jayanth2024eqnio} learns canonical displacement priors in a gravity-aligned frame for rotation consistency. TartanIMU \cite{zhao2025tartan} extends IO to a foundation model trained on 100+ hours of IMU data across cars, drones, robots, and humans, achieving cross-domain generalization through low-rank fine-tuning and online adaptation. M2EIT \cite{Li_2025_ICCV} further introduces a mixture-of-experts framework that fuses spatial, temporal, and frequency features for state-of-the-art robustness across motion domains.

Orthogonal to prior work, we introduce a human-pose-guided prior within a multi-stage framework, grounding motion estimation in body dynamics to improve both accuracy and robustness. We further leverage multi-sensor fusion on AR glasses—combining multiple IMUs and auxiliary sensors (barometer, magnetometer) already available on existing devices~\cite{engel2023aria}—to further mitigate long-term instability and bias drift.

\subsection{Body Motion Capture for IMU Devices}
Inertial measurement units (IMUs) have become an attractive sensing modality for motion capture due to their compact size, low power consumption, and portability. Commercial systems such as Xsens \cite{xsens} employ many high-grade IMUs (typically 17) for high-fidelity full-body tracking but remain intrusive and impractical for everyday use.

To improve practicality, recent research has explored reconstructing full-body pose from fewer sensors. SIP \cite{von2017sparse} demonstrated that six Xsens IMUs can reconstruct full-body motion using offline optimization. Deep Inertial Poser (DIP) \cite{huang2018deep} introduced a bi-directional LSTM model for pose estimation from six IMUs, while TransPose \cite{yi2021transpose} extended this to estimate global translation via a multi-stage architecture. TIP \cite{jiang2022transformer} applied Transformers for non-planar motion, and PIP \cite{yi2022physical}, PNP \cite{yi2024physical}, and GlobalPose \cite{yi2025improving} incorporated physics-based optimization for more plausible motion reconstruction.

Building on these advances, recent work leverages IMUs embedded in everyday devices. IMUPoser \cite{mollyn2023imuposer} demonstrated that arbitrary combinations of IMUs in earbuds, smartwatches, and smartphones can reconstruct full-body pose using masking strategies for missing sensors, while MobilePoser \cite{Xu2024MobilePoser} extended this approach with a multi-stage framework that jointly estimates pose and global translation. DiffusionPoser \cite{van2024diffusionposer} further employed a diffusion-based model to inpaint missing signals during denoising, supporting more flexible sensor placements.

These works pave the way for reconstructing full-body motion from sparse IMU configurations, but they often rely on multiple on-body IMUs. We propose learning a general human-pose prior from a single IMU which, when integrated into inertial odometry models, significantly reduces error and drift.

\subsection{Sensor Fusion for IMU Motion Tracking}
IMUs are compact and low-cost but are sensitive to noise, leading to drift over time \cite{zuo2025transformer}. To mitigate these limitations, extensive research has explored complementing IMUs with other sensors to improve stability and accuracy. The most common is visual-inertial odometry (VIO), which jointly optimizes motion estimates from IMU and camera observations. LiDAR–inertial odometry (LIO) follows a similar principle, using geometric depth constraints from LiDAR scans to improve robustness in low-light or textureless environments. We refer readers to comprehensive surveys for detailed reviews of VIO and LIO methods \cite{huang2019visualsurvey, lee2024lidar}. Also, MINS \cite{lee2023minsefficientrobustmultisensoraided} proposes a tightly coupled multisensor-aided inertial navigation system that fuses IMU, wheel encoders, cameras, LiDAR, and GNSS to address asynchronous measurements.

Cameras and LiDARs are power-intensive and often impractical for lightweight platforms such as AR glasses. Recent work instead incorporates non-visual modalities—magnetometers, barometers, ultrasound, and ToF sensors—for efficient motion tracking. MagShield \cite{shao2025magshieldbetterrobustnesssparse} uses magnetic cues to detect disturbances and correct orientation errors. BaroPoser \cite{Zhang_2025} exploits barometric pressure on wearables to recover global translation on non-flat terrain. UltraPoser \cite{li2025ultraposer} leverages ultrasound via built-in speakers and microphones to expand body-area sensing, while ToF-IP \cite{yao2025tof} integrates Time-of-Flight depth with sparse IMUs to constrain geometry and reduce drift. There is another line of work that fuses additional sensors into an EKF \cite{laidig2023vqf,madgwick2011estimation,mahony2008nonlinear}. However, little work has explored the fusion of easily accessible sensors, such as barometers and magnetometers, for neural inertial odometry.

\section{Method}

We formulate the inertial odometry (IO) problem as estimating a device trajectory, $T$, from a time series of IMU measurements comprising linear acceleration and angular velocity. Existing approaches differ in how they reconstruct motion from these signals. 

\begin{figure}
    \centering
    \includegraphics[width=1\linewidth]{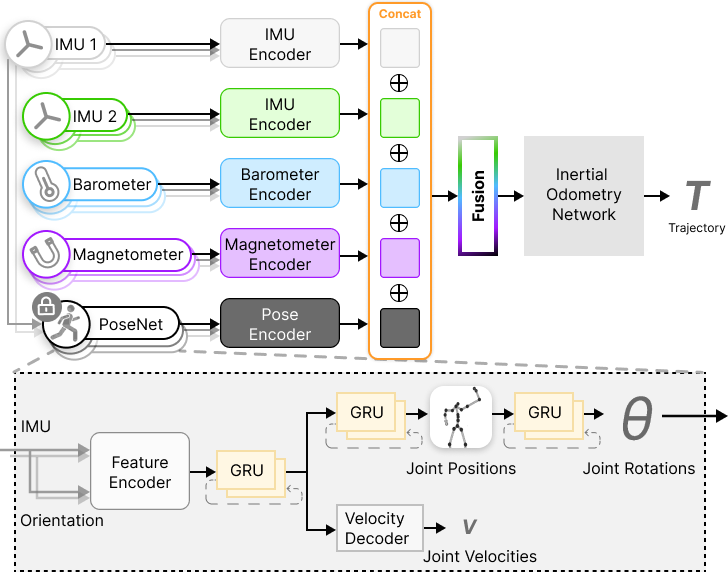}
    \caption{Overview of the MARIO inertial odometry framework. (a) We first learn a PoseNet that takes IMU measurements and orientations as input and predicts human pose (represented by joint rotations). (b) We encode multiple sensor signals—secondary IMU, barometer, magnetometer, and the pose estimated by the frozen PoseNet—concatenate their features through a fusion layer, and then feed the fused representation to an existing inertial odometry model to predict the trajectory.}
    \label{fig:placeholder}
\end{figure}

We aim to show that by explicitly introducing a human motion prior—learned from IMU signals—and integrating complementary sensing modalities, inertial tracking can be made more stable, accurate, and generalizable. As illustrated in Figure~\ref{fig:placeholder}, our framework consists of two main components:
(1) PoseNet, which learns a human motion prior from a single IMU to provide kinematic structure and temporal consistency; and
(2) a Multi-Sensor Fusion Module, which combines magnetometer, barometer, and secondary-IMU signals through a learned mid-level fusion strategy to enhance robustness and long-term stability.

To validate our approach, we integrate our modules into four representative IO architectures. AirIO \cite{qiu2025airiolearninginertialodometry} predicts body-frame velocity and transforms it to global coordinates using gyroscope-derived rotations before integration. TLIO \cite{Liu_2020} instead directly regresses 3D positional displacements over fixed 1-second windows using a ResNet architecture with gravity-aligned input. EqNIO\cite{jayanth2024eqnio} extends this displacement-based approach by learning in a canonical O(2) equivariant frame for improved rotation consistency. Finally, RoNIN-LSTM \cite{yan2019roninrobustneuralinertial} uses bi-directional LSTM for 2D velocity, we modified it to output 3D velocity vectors  for fair comparison.

\subsection{PoseNet}
We introduce PoseNet, a lightweight network that learns a pose prior from a single IMU stream. The network takes linear accelerations, angular velocities, and orientations as input and predicts body pose in SMPL format using a 6D rotation representation \cite{zhou2020continuityrotationrepresentationsneural}.

Following AirIO\cite{qiu2025airiolearninginertialodometry} archiecture to extract IMU feature information, PoseNet uses a hybrid convolutional–recurrent architecture designed to capture both short-term inertial dynamics and long-range temporal dependencies. IMU signals are first encoded by a temporal CNN that extracts local motion patterns, followed by stacked GRUs that sequentially predict joint positions and then rotations in the SMPL format. We estimate only nine joint angles corresponding to the pelvis and lower limbs, focusing on locomotion-related joints that dominate global motion while reducing model complexity. Estimating arm or hand motion from head-mounted IMUs is inherently underconstrained and thus excluded. Therefore, PoseNet will give a 9$\times$6D dimension pose vector each timestamp, which will be fed into our fusion module described in the next subsection.
We train the pose prior using an $\ell_2$ loss on joint positions and 6D joint rotations, and a Huber loss on joint-angle velocities to encourage temporal smoothness. Let $\mathbf{J}$ and $\hat{\mathbf{J}}$ denote the ground-truth and predicted joint positions, and $\boldsymbol{\theta}$ and $\hat{\boldsymbol{\theta}}$ the corresponding 6D rotations:

\[
\mathcal{L}_{\text{pos}}
\;=\;
 \left\| \hat{\mathbf{J}} - \mathbf{J} \right\|_2^2,
\qquad
\mathcal{L}_{\text{ang}}
\;=\;
\left\| \hat{\boldsymbol{\theta}} - \boldsymbol{\theta} \right\|_2^2.
\]
Given ground-truth velocity $v$ and predicted velocity $\hat{v}$, we compute the velocity loss defined by a Huber loss function:
\[
\mathcal{L}_{\text{vel}} =
\begin{cases}
\frac{1}{2}(\hat{v} - v)^2, & \text{if } |\hat{v} - v| < \delta, \\[6pt]
\delta \big(|\hat{v} - v| - \tfrac{1}{2}\delta \big), & \text{otherwise},
\end{cases}
\]
with $\delta = 0.005$. The total training objective is
\[
\mathcal{L} = 
\lambda_{\text{pos}}\mathcal{L}_{\text{pos}} +
\lambda_{\text{ang}}\mathcal{L}_{\text{ang}} +
\lambda_{\text{vel}}\mathcal{L}_{\text{vel}},
\]
where $\lambda_{\text{pos}} = 0.05$, $\lambda_{\text{ang}} = 0.05$, and $\lambda_{\text{vel}} = 1$.

PoseNet is trained on the Nymeria dataset \cite{ma2024nymeriamassivecollectionmultimodal}, which contains over 300 hours of synchronized IMU and ground-truth pose data collected from head-mounted devices across diverse activities and environments.

\begin{figure}
    \centering
    \includegraphics[width=1\linewidth]{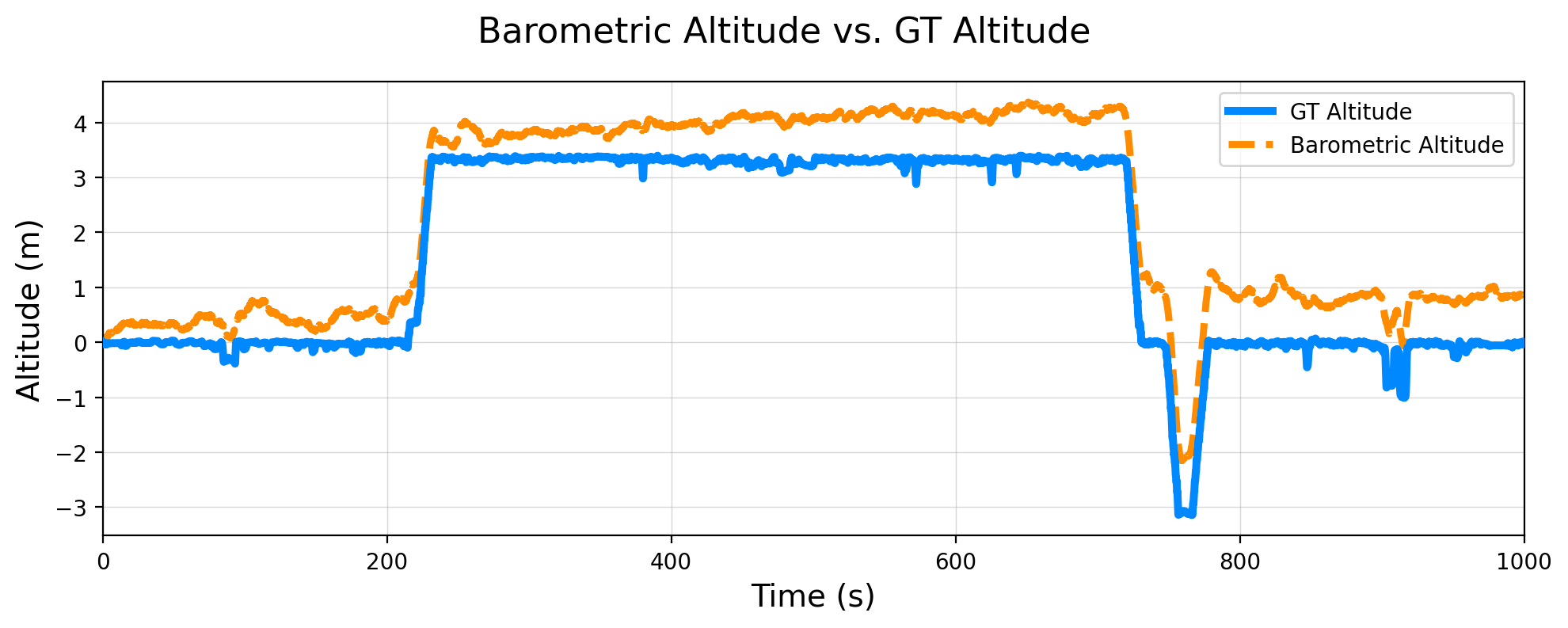}
    \caption{Visualization of altitude from the barometer compared with the ground-truth altitude. This demonstrates that the barometer provides valuable altitude information.}
    \label{baro_figure}
\end{figure}

\begin{figure}
    \centering
    \includegraphics[width=1\linewidth]{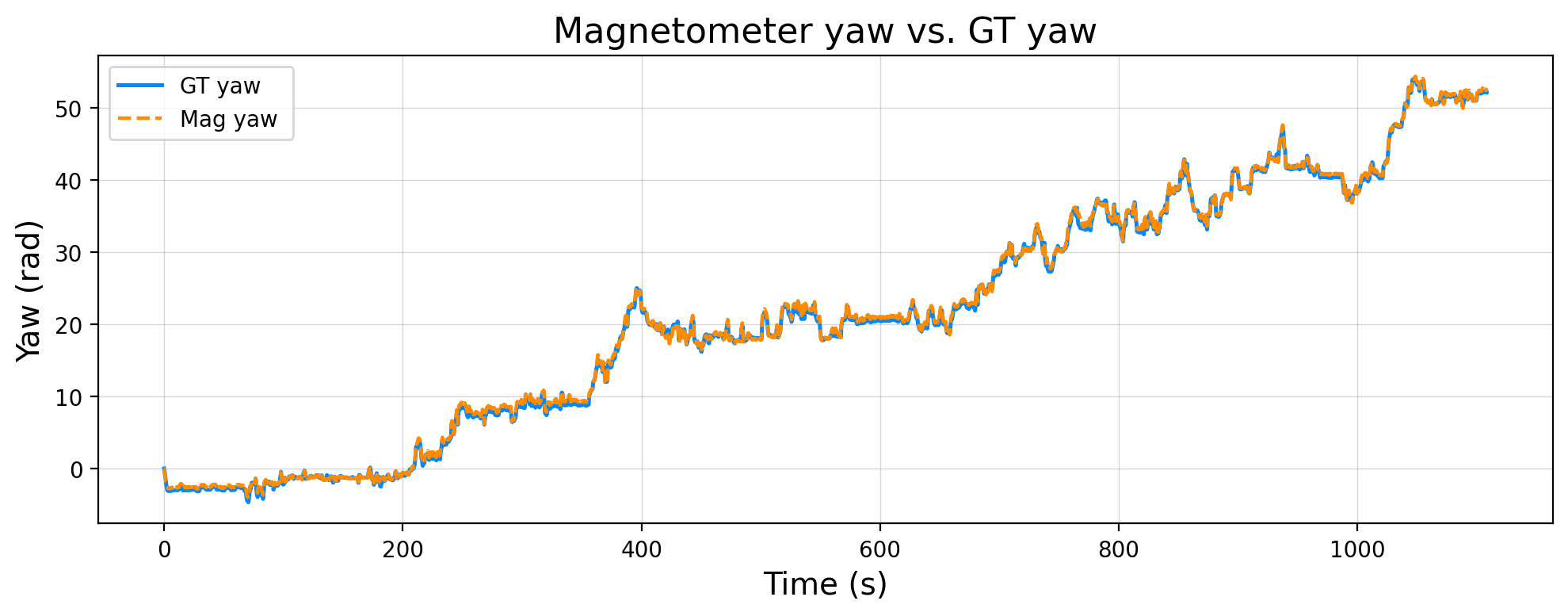}
    \caption{Visualization of magnetometer data compared with ground-truth yaw orientation. The magnetometer signal tracks the ground-truth, demonstrating its utility for constraining heading drift in inertial odometry.}
    \label{mag_figure}
\end{figure}

\subsection{Multi-Sensor Fusion Module}
Predicting motion from IMU signals is inherently prone to drift due to sensor noise and bias.  To address this limitation, we introduce a Multi-Sensor Fusion Module that combines complementary signals from a barometer, a magnetometer, and a secondary IMU that are already available on AR glasses \cite{engel2023aria}. These additional modalities supply grounding for elevation and heading and provide data redundancy that improves robustness and long-term stability.

\paragraph{Barometer}
A barometer measures ambient air pressure as a scalar time series in pascals. We convert pressure $p$ (Pa) to altitude $h$ (m) using the hypsometric relation under the International Standard Atmosphere (ISA) model, assuming sea-level pressure $P_0=101{,}325\,\text{Pa}$, temperature $T_0$, and lapse rate $L$:
\[
h \;\approx\; K\!\left(1 - \bigl(\tfrac{p}{P_0}\bigr)^{n}\right),
\qquad
K=\tfrac{T_0}{L}, \quad n=\tfrac{R\,L}{g_0},
\]
where $R$ is the specific gas constant for dry air and $g_0$ is standard gravity.  
The altitude signal is generally smooth but may drift slowly due to weather or indoor pressure changes. To reduce noise, we apply a moving-average filter to denoise both the altitude and its derivative vertical velocity, yielding a 1D velocity estimate at each timestamp that is later used in the fusion module. Figure~\ref{baro_figure} shows that barometric altitude closely follows ground-truth, indicating that pressure provides a strong cue for estimating altitude changes.

\paragraph{Magnetometer}
A magnetometer measures the local magnetic field as a 3D vector, which can be used to infer heading relative to magnetic north. Despite susceptibility to magnetic disturbances indoors, it provides a valuable absolute orientation cue when calibrated.

We compute the yaw angle by combining the magnetometer reading $\mathbf{m}$ with gravity $\mathbf{g}$ estimated from the IMU. After normalization,
$\hat{\mathbf{m}} = \mathbf{m} / \lVert \mathbf{m} \rVert$ and
$\hat{\mathbf{g}} = \mathbf{g} / \lVert \mathbf{g} \rVert$,
tilt compensation removes the vertical component:
\[
\mathbf{E} = \frac{\hat{\mathbf{m}} \times \hat{\mathbf{g}}}{\lVert \hat{\mathbf{m}} \times \hat{\mathbf{g}}\rVert}, 
\qquad
\mathbf{N} = \hat{\mathbf{g}} \times \mathbf{E},
\]
and the yaw is computed as
\[
\psi = \operatorname{atan2}\!\big((\mathbf{E})_z,\; (\mathbf{N})_z\big).
\]
This 1D yaw signal serves as the magnetometer input to the fusion module. 
Figure~\ref{mag_figure} compares the derived yaw with ground-truth, showing alignment and demonstrating that the magnetometer provides a useful heading cue.

\paragraph{Secondary IMU}
An accelerometer on a moving rigid body measures not only linear acceleration but also rotation-induced terms. For a body with linear acceleration $\mathbf{a}_0$, angular velocity $\boldsymbol{\omega}$, and angular acceleration $\boldsymbol{\alpha}$, an accelerometer at position $\mathbf{r}$ records
\[
\mathbf{a}
= \mathbf{a}_0
  + \boldsymbol{\alpha}\!\times\!\mathbf{r}
  + \boldsymbol{\omega}\!\times\!\bigl(\boldsymbol{\omega}\!\times\!\mathbf{r}\bigr)
  + \mathbf{b} + \mathbf{n},
\]
where $\mathbf{b}$ and $\mathbf{n}$ denote bias and noise.  
Adding a second, spatially separated IMU improves the observability of rotational motion and helps disentangle translational acceleration from rotation-induced effects. The dual-IMU configuration also provides redundancy that reduces the impact of sensor bias and noise. In the Aria glasses used in the Nymeria \cite{ma2024nymeriamassivecollectionmultimodal} and Aria Everyday Activities datasets \cite{lv2024ariaeverydayactivitiesdataset}, the primary IMU is located on the right temple and the secondary IMU on the left. Similar to the primary IMU, the secondary IMU provides a 6D (accelerometer + gyroscope) measurement vector at each timestamp for sensor fusion.

\subsection{Feature Fusion Strategy}
We use mid-level feature fusion tailored to the latent state of inertial odometry (IO). Each auxiliary stream (magnetometer, barometer, secondary IMU) is first time-aligned to the primary IMU. We then encode each stream with a small causal temporal CNN to obtain per-step features $f_{\text{mag}}(t)$, $f_{\text{baro}}(t)$, $f_{\text{imu2}}(t)$ and also for $f_{\text{pose}}(t)$. The primary IMU (accelerometer + gyroscope) is passed through the same architecture to yield $f_{\text{imu}}(t)$. At time $t$, we concatenate:
\begin{equation}
  f_{\text{fusion}}(t) = \big[\, 
    f_{\text{imu}}(t)\,;\, 
    f_{\text{mag}}(t)\,;\, 
    f_{\text{baro}}(t)\,;\, 
    f_{\text{imu2}}(t)\,;\, 
    f_{\text{pose}}(t) 
  \,\big].
\end{equation}
and feed $f_{\text{fusion}}(t)$ to a MLP fusion layer followed by existing inertial odometry models to estimate the motion state.

\section{Experiments}

\subsection{Dataset}

We evaluate our approach across three different datasets, Nymeria \cite{ma2024nymeriamassivecollectionmultimodal} , Aria Everyday Activities \cite{lv2024ariaeverydayactivitiesdataset}  and TLIO \cite{Liu_2020}. The Nymeria dataset serves as our primary training source, partitioned with an 80/10/10 split for train/validation/test. For additional evaluation on household activities, we use the Aria Everyday Activities (AEA) as a test dataset with a model pretrained on Nymeria. For baseline comparisons, we incorporate the TLIO dataset following the original repository's train/validation/test split~\cite{Liu_2020}.

\paragraph{Nymeria Dataset} 
The Nymeria dataset~\cite{ma2024nymeriamassivecollectionmultimodal} comprises 1200 sequences of 15-20 minute duration from 264 subjects in 50 varied indoor and outdoor environments, totaling 300 hours and 400km of movement. The data captures 20 different scenarios including athletic activities, meals, work tasks, outdoor excursions, and cycling. Project Aria glasses record synchronized multimodal signals: dual IMUs at 800Hz and 1kHz, magnetometer at 10Hz, and barometer at 50Hz. Ground-truth full-body pose is captured using an Xsens MVN Link suit with 17 inertial units at 240 Hz. We convert these Xsens poses into the SMPL format to supervise PoseNet. The Project Aria MPS system provides the 3D trajectories at 1 kHz used as ground-truth for inertial odometry.

\paragraph{Aria Everyday Activities Dataset} 
The AEA dataset~\cite{lv2024ariaeverydayactivitiesdataset} contains 143 recording sessions across five distinct indoor environments, totaling 7.3 hours. Activities include typical household routines: meal preparation, tidying, eating, and social interaction. Recording setup and ground-truth systems match specifications from Nymeria dataset.

\paragraph{TLIO Dataset} 
The TLIO dataset~\cite{Liu_2020} features approximately 60 hours of walking motion recorded via head-worn hardware with a Bosch BMI055 inertial sensor. The collection focuses on pedestrian movement patterns within interior spaces. Unlike Nymeria and AEA, TLIO provides only IMU measurements without additional sensing modalities, with ground-truth trajectories derived from Visual Inertial Odometry.

\subsection{Implementation Details}
We conduct all experiments on NVIDIA GeForce RTX 4090 and A40 GPUs. AirIO uses a sliding window of 1,000 samples (stride 100), while EqNIO, TLIO, and RoNIN-LSTM use 200 samples with stride 10 for TLIO/Aria Everyday datasets and stride 100 for Nymeria. All signals are resampled to 200 Hz; barometer and magnetometer readings are interpolated to match the IMU rate. Unless otherwise specified, models are trained on Nymeria's training split, evaluated on its test split, then assessed on Aria Everyday for cross-dataset generalization. The pose prior is trained exclusively on Nymeria and frozen during training and evaluation.

We use a learning rate of $5\times 10^{-4}$ for AirIO and $1\times 10^{-4}$ for TLIO. During training, we condition the models on ground-truth orientations, whereas at test time we use orientations inferred from the gyroscope. For TLIO, EqNIO, and RoNIN-LSTM, we train the covariance branch from the first iteration, which we find leads to more stable convergence on Nymeria. For gravity compensation in the raw IMU accelerometers, we subtract gravity using the known device orientation.

\subsection{Evaluation Metrics}
Performance is reported using Absolute Trajectory Error (ATE), Relative Translational Error at 1\,s and 5\,s (RTE-1s / RTE-5s), and a segment-wise drift rate (``Drifting'').
ATE is computed as the root-mean-square error between estimated and ground-truth positions after temporal synchronization and rigid alignment:
\begin{equation}
\mathrm{ATE} \;=\; \sqrt{\frac{1}{n}\sum_{i=1}^{n} \big\| \mathbf{p}_i - \hat{\mathbf{p}}_i \big\|^{2}} \, .
\end{equation}
To better analyze ATE, we further decompose it into horizontal and vertical components, where the horizontal error is measured in 2D, capturing motion on the ground plane, and the vertical error is measured in 1D, capturing only upward and downward motion.
RTE-$\Delta t$ measures the mean error of relative displacements over sliding windows of duration $\Delta t \in \{1,5\}\,\mathrm{s}$:
\begin{equation}
\mathrm{RTE}_{\Delta t} \;=\; \frac{1}{n}\sum_{i=1}^{n}
\left\|
\big(\mathbf{p}_{i+\Delta t}-\mathbf{p}_{i}\big)
-
\big(\hat{\mathbf{p}}_{i+\Delta t}-\hat{\mathbf{p}}_{i}\big)
\right\| .
\end{equation}
\noindent
Drifting is reported as the mean relative error (percentage) with respect to the ground-truth displacement over the same windows.

\subsection{Results}

We present results on the Nymeria and Aria datasets, integrating our PoseNet and
Multi-Sensor Fusion Module into four IO architectures: AirIO \cite{qiu2025airiolearninginertialodometry}, TLIO \cite{Liu_2020}, EqNIO \cite{jayanth2024eqnio}, and
RoNIN-LSTM \cite{yan2019roninrobustneuralinertial}. We use (+Pose) to denote models with our pose prior and (+All) for
models using both pose and all sensor modalities.

\begin{table*}[t]
  \centering
  \caption{Nymeria dataset results. Metrics: ATE (m; horizontal H, vertical V), RTE-5s (m), RTE-1s (m), Drift (\%). Lower is better. The best result within each group (\textbf{Single IMU} and \textbf{Multi-Sensor}) is shown in bold.}
  \label{tab:nymeria_cpf_nymeria}
  \setlength{\tabcolsep}{4pt}
  \renewcommand{\arraystretch}{1.1}
  \footnotesize

  \begin{tabular}{ll
      >{}c
      >{\columncolor{lightorange}}c|
      c c c
      >{\columncolor{lightorange}}c}
    \toprule
    & & \multicolumn{2}{c|}{Single IMU} & \multicolumn{4}{c}{Multi-Sensor} \\
    \midrule
    Model & Metric & Base & +Pose & +sec. IMU & +Baro & +Mag & +All \\
    \midrule

    % ------------------ AirIO ------------------
    \multirow{4}{*}{AirIO} 
    & ATE (H,V) & 6.85 (6.43, 1.95)
      & \textbf{5.22 (4.89, 1.43)}
      & 5.20 (4.81, 1.63)
      & 5.25 (5.08, 1.00)
      & 5.71 (5.30, 1.71)
      & \textbf{4.64 (4.30, 1.36)} \\
    & RTE-5s & 0.363
      & \textbf{0.323}
      & 0.332
      & 0.362
      & 0.360
      & \textbf{0.297} \\
    & RTE-1s & 0.099
      & \textbf{0.085}
      & 0.088
      & 0.109
      & 0.098
      & \textbf{0.078} \\
    & Drift & 3.56
      & \textbf{2.35}
      & 2.52
      & 2.56
      & 2.77
      & \textbf{2.16} \\
    \midrule

    % ------------------ TLIO ------------------
    \multirow{4}{*}{TLIO}
    & ATE (H,V) & 10.19 (9.87, 1.09)
      & \textbf{7.97 (7.52, 1.46)}
      & 6.45 (6.09, 0.93)
      & 5.85 (5.54, 0.81)
      & 8.57 (8.17, 1.45)
      & \textbf{5.73 (5.46, 0.77)} \\
    & RTE-5s & 0.322
      & \textbf{0.281}
      & 0.300
      & 0.260
      & 0.304
      & \textbf{0.242} \\
    & RTE-1s & 0.103
      & \textbf{0.096}
      & 0.097
      & 0.088
      & 0.099
      & \textbf{0.083} \\
    & Drift & 6.46
      & \textbf{4.94}
      & 3.70
      & 3.34
      & 5.17
      & \textbf{3.64} \\
    \midrule

    % ------------------ EqNIO ------------------
    \multirow{4}{*}{EqNIO}
    & ATE (H,V) & \textbf{7.63 (7.21, 1.09)}
      & 7.65 (7.26, 1.08)
      & 5.49 (5.21, 0.70)
      & 5.27 (5.02, 0.62)
      & 6.85 (6.48, 0.86)
      & \textbf{4.52 (4.18, 0.96)} \\
    & RTE-5s & 0.379
      & \textbf{0.316}
      & 0.273
      & 0.270
      & 0.331
      & \textbf{0.231} \\
    & RTE-1s & 0.122
      & \textbf{0.101}
      & 0.091
      & 0.093
      & 0.107
      & \textbf{0.081} \\
    & Drift & \textbf{3.93}
      & 4.30
      & 2.84
      & 2.85
      & 3.39
      & \textbf{2.40} \\
    \midrule

    % ------------------ RoNIN-LSTM ------------------
    \multirow{4}{*}{RoNIN-LSTM}
    & ATE (H,V) & 9.10 (8.77, 0.97)
      & \textbf{6.83 (6.51, 0.92)}
      & 8.76 (8.34, 1.30)
      & \textbf{5.88 (5.59, 0.76)}
      & 5.93 (5.60, 0.95)
      & 5.89 (5.66, 0.64) \\
    & RTE-5s & 0.330
      & \textbf{0.281}
      & 0.289
      & 0.261
      & 0.254
      & \textbf{0.246} \\
    & RTE-1s & 0.103
      & \textbf{0.092}
      & 0.093
      & 0.088
      & 0.085
      & \textbf{0.084} \\
    & Drift & 6.22
      & \textbf{3.99}
      & 6.13
      & \textbf{3.44}
      & 3.64
      & 3.61 \\
    \bottomrule
  \end{tabular}
\end{table*}

\begin{table*}[t]
  \centering
  \caption{Aria dataset results using pretrained model from Nymeria dataset. Metrics: ATE (m; horizontal H, vertical V), RTE-5s (m), RTE-1s (m), Drift (\%). Lower is better. The best result within each group (\textbf{Single IMU} and \textbf{Multi-Sensor}) is shown in bold.}
  \label{tab:aria_cpf_nymeria}
  \setlength{\tabcolsep}{4pt}
  \renewcommand{\arraystretch}{1.1}
  \footnotesize
  \begin{tabular}{ll
      >{}c
      >{\columncolor{lightorange}}c|
      c c c
      >{\columncolor{lightorange}}c}
    \toprule
    & & \multicolumn{2}{c|}{Single IMU} & \multicolumn{4}{c}{Multi-Sensor} \\
    \midrule
    Model & Metric & Base & +Pose & +sec. IMU & +Baro & +Mag & +All \\
    \midrule

    \multirow{4}{*}{AirIO} 
    & ATE (H,V) & 1.25 (1.18, 0.29) & \textbf{0.86 (0.79, 0.25)} & 1.18 (0.92, 0.66) & 1.03 (0.99, 0.22) & 1.29 (1.21, 0.33) & \textbf{0.89 (0.82, 0.28)} \\
    & RTE-5s & 0.334 & \textbf{0.218} & 0.227 & 0.273 & 0.300 & \textbf{0.201} \\
    & RTE-1s & 0.106 & \textbf{0.062} & 0.068 & 0.089 & 0.099 & \textbf{0.062} \\
    & Drift & 11.38 & \textbf{5.32} & 7.35 & 8.55 & 9.48 & \textbf{6.04} \\
    \midrule

    \multirow{4}{*}{TLIO}
    & ATE (H,V) & 1.51 (1.45, 0.21) & \textbf{1.23 (1.13, 0.26)} & 1.05 (0.95, 0.25) & \textbf{1.00 (0.93, 0.18)} & 1.12 (1.00, 0.29) & 1.02 (0.96, 0.17) \\
    & RTE-5s & 0.246 & \textbf{0.222} & 0.231 & 0.202 & 0.231 & \textbf{0.195} \\
    & RTE-1s & 0.081 & \textbf{0.075} & 0.077 & 0.070 & 0.078 & \textbf{0.069} \\
    & Drift & 9.69 & \textbf{8.09} & 7.76 & \textbf{6.68} & 8.06 & 7.67 \\
    \midrule

    \multirow{4}{*}{EqNIO}
    & ATE (H,V) & \textbf{1.20 (1.11, 0.24)} & 1.24 (1.17, 0.20) & 0.92 (0.85, 0.18) & 1.03 (0.98, 0.16) & 1.08 (1.01, 0.20) & \textbf{0.89 (0.84, 0.13)} \\
    & RTE-5s & 0.270 & \textbf{0.233} & 0.203 & 0.212 & 0.246 & \textbf{0.186} \\
    & RTE-1s & 0.093 & \textbf{0.079} & 0.072 & 0.075 & 0.084 & \textbf{0.067} \\
    & Drift & \textbf{8.18} & 8.45 & \textbf{6.43} & 8.14 & 7.28 & 6.67 \\
    \midrule

    \multirow{4}{*}{RoNIN-LSTM}
    & ATE (H,V) & 1.39 (1.31, 0.20) & \textbf{1.04 (0.97, 0.19)} & 1.45 (1.31, 0.31) & 0.99 (0.92, 0.17) & \textbf{0.95 (0.88, 0.18)} & 1.05 (0.99, 0.14) \\
    & RTE-5s & 0.270 & \textbf{0.212} & 0.234 & 0.206 & \textbf{0.196} & 0.201 \\
    & RTE-1s & 0.085 & \textbf{0.073} & 0.077 & 0.071 & \textbf{0.069} & 0.070 \\
    & Drift & 13.22 & \textbf{7.60} & 12.51 & 7.74 & \textbf{7.38} & 7.49 \\
    \bottomrule
  \end{tabular}
\end{table*}

\begin{figure}[h]
    \centering
    \includegraphics[width=1\linewidth]{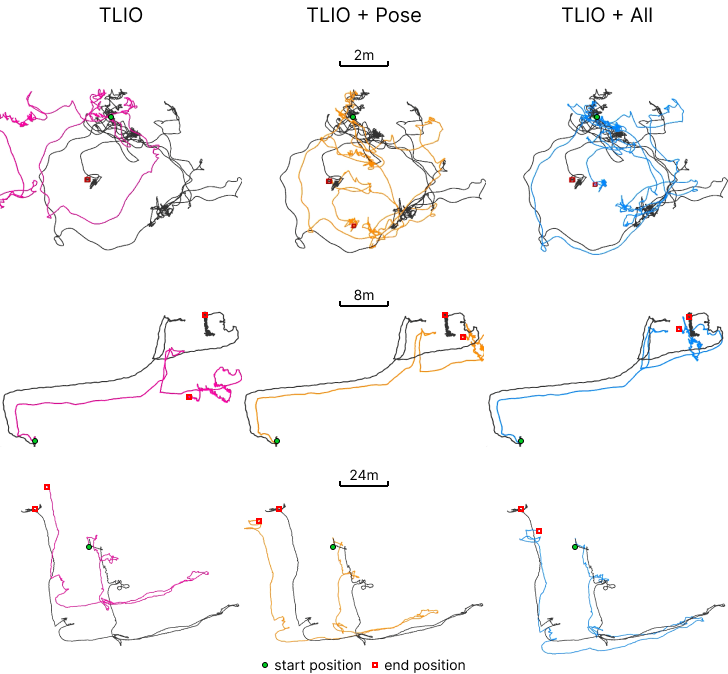}
    \caption{Trajectory visualizations of TLIO, TLIO+Pose, and TLIO+All on Nymeria dataset. TLIO+Pose improves over TLIO, and TLIO+All further reduces drift and tracking errors.}
    \label{sample_viz_figure}
\end{figure}

Table~\ref{tab:nymeria_cpf_nymeria} and~\ref{tab:aria_cpf_nymeria} present our main results on Nymeria and Aria Everyday datasets. On Nymeria, our full system (+All) achieves consistent improvements across all architectures: AirIO shows 32\% ATE reduction (6.85m → 4.64m) and 39\% drift reduction (3.56\% → 2.16\%), while TLIO demonstrates even larger gains with 44\% ATE improvement (10.19m → 5.73m) and 44\% drift reduction (6.46\% → 3.64\%). EqNIO and RoNIN-LSTM follow similar trends with 41\% and 35\% ATE reductions respectively. Adding PoseNet alone delivers substantial benefits, reducing RTE-5s by 11\% for AirIO and 13\% for TLIO. Individual sensor modalities contribute complementary cues, with the barometer particularly effective for vertical stability. On Aria (Table~\ref{tab:aria_cpf_nymeria}), models pretrained on Nymeria generalize effectively without fine-tuning, maintaining strong performance with 29-32\% ATE improvements. The pose prior shows robust cross-domain transfer, reducing drift by up to 53\% on Aria. Figure~\ref{sample_viz_figure} shows this progressive improvement with three sample sequences across TLIO variants. Furthermore, our AirIO model with full sensor fusion and pose prior (+All) achieves 133.6M FLOPs per inference and 315 FPS on NVIDIA A40 GPU, enabling real-time deployment. 

\begin{figure}[t]
    \centering
    \includegraphics[width=1\linewidth]{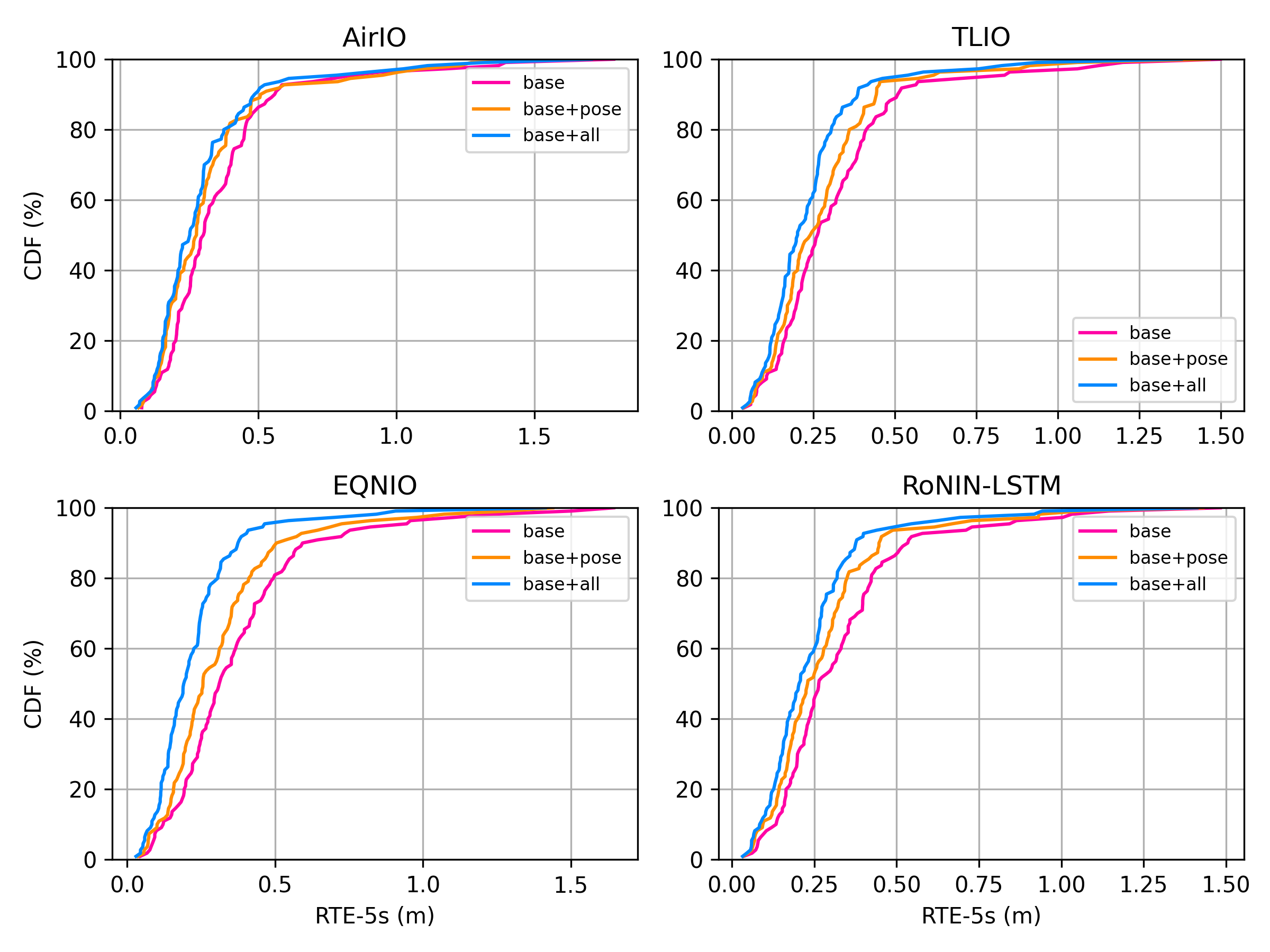}
    \caption{RTE-5s CDF on the Nymeria dataset for AirIO, TLIO, EqNIO, and RoNIN-LSTM. We show the cumulative distribution of 5-second relative trajectory error for all four base models.}
    \label{rte_cdf_figure}
\end{figure}

In Figure~\ref{rte_cdf_figure}, we demonstrate the cumulative distribution function (CDF) of RTE-5s for AirIO, TLIO, EqNIO, and RoNIN-LSTM on the Nymeria dataset. Adding the pose prior consistently lowers RTE-5s for all models, and incorporating all sensors together with the pose prior further reduces RTE-5s across the entire Nymeria test set.

\section{Ablation Studies}

\begin{table}[t]
  \centering
  \caption{TLIO dataset result. Metrics reported: ATE (m; horizontal H, vertical V), RTE-5s (m), RTE-1s (m), and \% Drifting (lower is better). RoNIN refers to RoNIN-LSTM variant.}
  \label{tab:pose_ablation}
  \resizebox{\columnwidth}{!}{%
    \begin{tabular}{lcccc}
      \toprule
      Model & ATE (H, V) & RTE-5s & RTE-1s & Drifting \\
      \midrule
      AirIO & 2.191 (2.034, 0.632) & 0.358 & 0.092 & 2.36 \\
      AirIO + Pose  & \textbf{1.950 (1.848, 0.462)} & \textbf{0.344} & \textbf{0.085} & \textbf{1.83} \\
      \midrule
      TLIO & 2.415 (2.259, 0.357) & 0.314 & 0.100 & 3.28 \\
      TLIO + Pose &  \textbf{2.023 (1.890, 0.249)} & \textbf{0.307} & \textbf{0.097} & \textbf{2.70} \\
      \midrule
      EQNIO & 2.116 (1.965, 0.319) & 0.339 & 0.104& \textbf{2.55} \\
      EQNIO + Pose & \textbf{2.073 (1.916, 0.315)} & \textbf{0.326}	& \textbf{0.100}	& 2.56 \\
      \midrule
      RoNIN & 2.953 (2.806, \textbf{0.335})&	0.345&	0.107&	4.06 \\
      RoNIN + Pose & \textbf{2.630} (\textbf{2.466}, 0.384)& \textbf{0.330} & \textbf{0.101}	& \textbf{3.76} \\
      \bottomrule
    \end{tabular}%
  }
\end{table}

\paragraph{Pose Prior Ablation} 
In Table~\ref{tab:pose_ablation}, we evaluate the pose prior's effectiveness 
on the TLIO dataset. We evaluate four baseline models—AirIO, TLIO, EqNIO, and RoNIN-LSTM—using PoseNet pretrained on the Nymeria dataset. The results show that incorporating PoseNet consistently improves ATE by 2–16\% across all models, indicating that our pose prior generalizes well across different datasets.

\paragraph{Gravity Removal Ablation} 

To investigate whether the model benefits from gravity-removed linear acceleration versus raw acceleration, we conduct an ablation study in Table~\ref{tab:gravity_ablation}. We compare models trained with (w/ g) and without (w/o g) gravitational components in the acceleration data. Results show that removing gravity consistently improves performance across both AirIO and TLIO with all sensors fused. For AirIO, gravity removal reduces ATE from 6.121m to 4.641m and drifting from 3.58\% to 2.16\%. Similarly, TLIO shows improvements with ATE decreasing from 6.613m to 5.734m when gravity is removed. These results suggest that linear acceleration is more informative than raw acceleration for head tracking tasks.

\begin{table}[t]
  \centering
  \caption{Nymeria results with “(+ all)” ablations (with/without gravity). Metrics: ATE (m), RTE-5s (m), RTE-1s (m), and \% Drifting (lower is better).}
  \label{tab:gravity_ablation}
  \setlength{\tabcolsep}{6pt}
  \renewcommand{\arraystretch}{1.15}
  \resizebox{\columnwidth}{!}{%
    \begin{tabular}{lccccc}
      \toprule
      Model & Gravity & ATE & RTE-5s & RTE-1s & Drifting \\
      \midrule
      AirIO + all & w/ g   & 6.121 & 0.347 & 0.110 &  3.58  \\
      AirIO + all & w/o g  & \textbf{4.641}  & \textbf{0.297} & \textbf{0.078} & \textbf{2.16} \\
      \midrule
      TLIO + all  & w/ g   & 6.613  & 0.287 & 0.095 & 4.01 \\
      TLIO + all  & w/o g  & \textbf{5.734}  & \textbf{0.242} & \textbf{0.083} & \textbf{3.64} \\
      \bottomrule
    \end{tabular}%
  }
\end{table}

\paragraph{Fusion Strategy Ablation} Table~\ref{tab:fusion_ablation} compares our sensor-fusion strategy with alternative fusion modalities. We evaluate (i) direct early-stage concatenation, where pose joint parameters are concatenated with accelerometer and gyroscope measurements, and (ii) cross-attention, where pose parameters are first encoded by a CNN-based pose encoder and the resulting pose features attend to all IMU features via a cross-attention module. The experiments indicate that our approach achieves better ATE, RTE, and drift performance in most cases.

\begin{table}[t]
  \centering
  \caption{Nymeria results with fusion strategy ablations. Metrics: ATE (m), RTE-5s (m), RTE-1s (m), and \% Drifting.}
  \label{tab:fusion_ablation}
  \setlength{\tabcolsep}{6pt}
  \renewcommand{\arraystretch}{1.15}
  \resizebox{\columnwidth}{!}{%
    \begin{tabular}{lccccc}
      \toprule
      Model & Fusion & ATE & RTE-5s & RTE-1s & Drifting \\
      \midrule
      AirIO + Pose & raw concat & 9.739 & 0.509 & 0.171 & 6.08 \\
      AirIO + Pose & cross attn & 5.767 & 0.398 & 0.116 & 2.64 \\
      AirIO + Pose & ours & \textbf{5.218} & \textbf{0.323} & \textbf{0.085} & \textbf{2.35} \\
      \midrule
      TLIO + Pose & raw concat & 8.323 & 0.282 & \textbf{0.092} & 5.18 \\
      TLIO + Pose & cross attn & 8.471 & 0.284 & 0.093 & 5.52 \\
      TLIO + Pose & ours & \textbf{7.972} & \textbf{0.281} & 0.096 & \textbf{4.94} \\
      \bottomrule
    \end{tabular}%
  }
\end{table}

\section{Limitations}

While the MARIO framework shows clear benefits from integrating human-pose priors and multi-sensor fusion for inertial odometry, several limitations remain. PoseNet introduces additional latency; future work could explore injecting kinematic priors directly into existing inertial odometry models for better efficiency. Magnetometers remain vulnerable to indoor disturbances, and barometric altitude is sensitive to pressure drift and weather changes. Future work includes learned reliability gating for corrupted sensors, self-calibration for misalignment, and further optimization for real-time deployment on resource-constrained wearables.
\section{Conclusion}

We present MARIO, an inertial odometry framework that augments off-the-shelf IO models with an IMU-inferred pose prior to ground motion estimation in human kinematics. Beyond standard IMU input, we also incorporate lightweight sensing modalities, including a barometer, a magnetometer, and a second on-glasses IMU, through an efficient multimodal fusion strategy. MARIO plugs into existing IMU-based backbones and improves trajectory accuracy by up to 42\%.

\clearpage
\setcounter{page}{1}
\maketitlesupplementary

\section{PoseNet Evaluation Experiment}
To evaluate the quality of our PoseNet, we compare it with Ego4o~\cite{wang2025ego4oegocentrichumanmotion}, including Ego4o-IMU, which uses head IMU only, and Ego4o, which additionally uses egocentric RGB images and motion descriptions. We report MPJPE and PA-MPJPE on the Nymeria dataset; since Ego4o uses its own curated splits, these results are not directly comparable and are included only to contextualize the scale of pose accuracy achievable with richer modalities. As shown in Table~\ref{tab:ego4o_results}, our method outperforms Ego4o-IMU and achieves comparable accuracy to Ego4o, despite using fewer modalities.

\begin{table}[ht]
    \centering
    \begin{tabular}{lcc}
        \hline
        Method & MPJPE & PA-MPJPE \\
        \hline
        \textcolor{lighttextgray}{Ego4o-IMU} & \textcolor{lighttextgray}{123.80} & \textcolor{lighttextgray}{85.45} \\
        \textcolor{lighttextgray}{Ego4o} & \textcolor{lighttextgray}{99.68} & \textcolor{lighttextgray}{72.60} \\
        \midrule
        PoseNet (ours) & 101.24 & 78.71 \\
        \hline
    \end{tabular}
    \caption{Comparison of our PoseNet, Ego4o-IMU~\cite{wang2025ego4oegocentrichumanmotion} and Ego4o~\cite{wang2025ego4oegocentrichumanmotion} on MPJPE and PA-MPJPE.}
    \label{tab:ego4o_results}
\end{table}

To further demonstrate that the pose prior is dataset-agnostic, we also train our PoseNet on the AMASS dataset and evaluate it on the Nymeria dataset. We observe that PoseNet trained on AMASS improves the performance of both the AirIO and TLIO models.

\begin{table}[t]
  \centering
  \caption{Results on the Nymeria dataset. Pose$_N$ and Pose$_A$ denotes PoseNet trained on the Nymeria and AMASS training sets, respectively}
  \label{tab:posenet_amass}
  \resizebox{\columnwidth}{!}{
    \begin{tabular}{lcccc}
      \toprule
      Model & ATE(m) & RTE-5s (m)& RTE-1s (m)& Drifting(\%)  \\
      \midrule
      AirIO & 6.85 & 0.363 & 0.099 & 3.56 \\
      AirIO + Pose$_A$ & 5.61 & 0.315 & 0.094 & 3.48 \\
      AirIO + Pose$_N$ & 5.22 & 0.323 & 0.085 & 2.35 \\
      \midrule
      TLIO & 10.19 & 0.322 & 0.103 & 6.46 \\
      TLIO + Pose$_A$ & 7.82 & 0.309 & 0.099 & 4.69 \\
      TLIO + Pose$_N$ & 7.97 & 0.281 & 0.096 & 4.94 \\
      \bottomrule
    \end{tabular}
  }
\end{table}

\section{Coordinate Details}

We convert all recordings to a Central Pupil Frame (CPF) for consistent training and evaluation. Raw SLAM data provides positions and orientations in an arbitrary world frame, while IMU measurements and velocities are in the device's body frame. This mismatch prevents proper velocity integration during training. We resolve this by choosing a world frame that is expressed in CPF (+X left, +Y up, +Z forward). 

CPF (Central Pupil Frame) is a rigid device-centric coordinate system with +X left, +Y up, +Z forward. We describe how raw device data is transformed into the CPF coordinate frame. At a chosen reference time, we (1) project the device’s CPF forward direction onto the horizontal plane to define the forward axis of the new world frame, and (2) align the vertical axis with gravity. This creates a unified, gravity-aligned frame that supports geometrically consistent transformations between the device's body frame and the world frame.

\textbf{Stage 1: Device \(\rightarrow\) CPF.} Using Aria calibration:
\begin{align}
\mathbf{v}_{\mathrm{CPF}} &= R_{\mathrm{CPF}\leftarrow\mathrm{device}}\;\mathbf{v}_{\mathrm{device}},
\label{eq:vectors-device-to-cpf-2col} \\
R_{\mathrm{world}\leftarrow\mathrm{CPF}}(t) &= 
R_{\mathrm{world}\leftarrow\mathrm{device}}(t)\;R_{\mathrm{device}\leftarrow\mathrm{CPF}},
\label{eq:poses1-device-to-cpf-2col} \\
R_{\mathrm{device}\leftarrow\mathrm{CPF}} &= 
R_{\mathrm{CPF}\leftarrow\mathrm{device}}^{\top}.
\label{eq:poses2-device-to-cpf-2col}
\end{align}

\textbf{Stage 2: CPF \(\rightarrow\) CPF-World (gravity-aligned, consistent heading).}
Choose a reference time \(t_0\) (30\,s after sequence start to allow SLAM to settle). Compute the CPF forward direction in world coordinates:
\begin{equation}
\mathbf{z}^{\mathrm{world}}_{\mathrm{CPF}}
= R_{\mathrm{world}\leftarrow\mathrm{CPF}}(t_0)\,[0,\,0,\,1]^{\top}.
\label{eq:zcpf-world}
\end{equation}
Project to the horizontal plane and normalize:
\begin{equation}
\hat{\mathbf{z}}_{\mathrm{hor}} =
\frac{[\,z_x,\, z_y,\, 0\,]^{\top}}{\lVert [\,z_x,\, z_y,\, 0\,] \rVert},
\quad (z_x, z_y, \cdot) = \mathbf{z}^{\mathrm{world}}_{\mathrm{CPF}}.
\label{eq:zhorizontal}
\end{equation}
Define CPF-world axes (expressed in world coordinates):
\begin{equation}
\begin{aligned}
\mathbf{y}_{\mathrm{cpfw}} &= [0,\,0,\,1]^{\top}, \\
\mathbf{z}_{\mathrm{cpfw}} &= \hat{\mathbf{z}}_{\mathrm{hor}}, \\
\mathbf{x}_{\mathrm{cpfw}} &= \mathbf{y}_{\mathrm{cpfw}} \times \mathbf{z}_{\mathrm{cpfw}}.
\end{aligned}
\label{eq:cpfw-axes}
\end{equation}
Assemble the rotation with columns as the axes in \eqref{eq:cpfw-axes}:
\begin{equation}
R_{\mathrm{world}\leftarrow\mathrm{cpfw}} =
\big[\,\mathbf{x}_{\mathrm{cpfw}}\;\; \mathbf{y}_{\mathrm{cpfw}}\;\; \mathbf{z}_{\mathrm{cpfw}}\,\big].
\label{eq:R-world-cpfw}
\end{equation}
Transform positions by rotation and centering at \(\mathbf{p}_{\mathrm{ref}}\):
\begin{equation}
\mathbf{p}_{\mathrm{cpfw}}(t) =
R_{\mathrm{world}\leftarrow\mathrm{cpfw}}^{\top}
\big(\mathbf{p}_{\mathrm{world}}(t) - \mathbf{p}_{\mathrm{ref}}\big).
\label{eq:position-cpfw}
\end{equation}

% \section{Additional Trajectory Visualization}
% In Figure~\ref{4_models}, we visualize the predicted trajectories of the base model, the base model + pose, and the base model + all for four different methods—AirIO, TLIO, EqNIO, and RoNIN-LSTM—on the Nymeria dataset. We observe that, for each model, +pose visibly reduces drift relative to the base model, and +all further aligns the trajectory around ground truth.

% \begin{figure*}
%     \centering
%     \includegraphics[width=1\linewidth]{figures/viz.png}
%     \caption{We show the trajectory predictions alongside the ground truth for AirIO, TLIO, EqNIO, and RoNIN-LSTM on one sequence from the Nymeria dataset.}
%     \label{4_models}
% \end{figure*}

\section{Additional Trajectory Visualization}
\begin{figure*}[!t]
    \centering
    \includegraphics[width=0.98\textwidth]{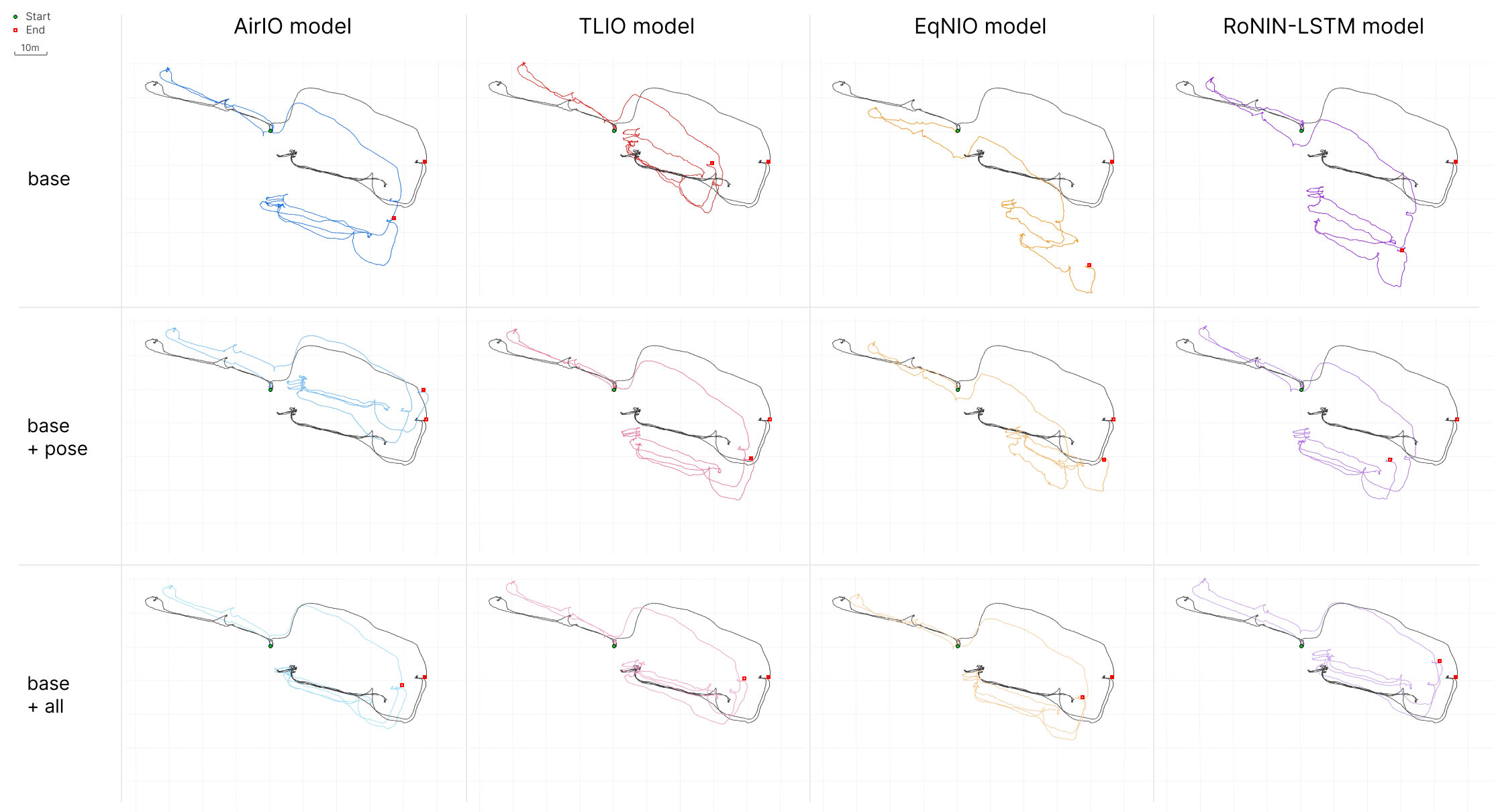}
    \caption{We show the trajectory predictions alongside the ground truth for AirIO, TLIO, EqNIO, and RoNIN-LSTM on one sequence from the Nymeria dataset.}
    \label{4_models}
\end{figure*}

In Figure~\ref{4_models}, we visualize the predicted trajectories of the base model, the base model + pose, and the base model + all for four different methods—AirIO, TLIO, EqNIO, and RoNIN-LSTM—on the Nymeria dataset. We observe that, for each model, +pose visibly reduces drift relative to the base model, and +all further aligns the trajectory around ground truth.

\FloatBarrier
{
    \small
    \bibliographystyle{ieeenat_fullname}
    \bibliography{main}
}

\end{document}